\documentclass[letterpaper, 10 pt, conference]{ieeeconf}  

\IEEEoverridecommandlockouts                              

\usepackage{graphicx} 
\usepackage[nolist]{acronym} 
\usepackage{amsmath}
\usepackage{amssymb}
\usepackage{balance}
\usepackage{siunitx}
\usepackage{bm}
\usepackage{booktabs} 
\usepackage{pgfplots}
\pgfplotsset{compat=newest}
\usepgfplotslibrary{groupplots}
\usepackage{silence}
\usepackage[caption=false,font=footnotesize]{subfig}
\usepackage{url}
\let\labelindent\relax
\usepackage[inline]{enumitem}
\title{\LARGE \bf
Behavioral Cloning for Robotic Connector Assembly: \\ An Empirical Study
}

\author{Andreas Kernbach$^{1,2}$, Daniel Bargmann$^{1}$, Werner Kraus$^{1}$ and Marco F. Huber$^{1,2}$
\thanks{$^{1}$The authors are with the Fraunhofer Institute for Manufacturing Engineering and Automation IPA, Stuttgart, Germany {\tt\small \{andreas.kernbach, daniel.bargmann, werner.kraus, marco.huber\}@ipa.fraunhofer.de}}%
\thanks{$^{2}$Institute of Industrial Manufacturing and Management IFF, University of Stuttgart, Germany {\tt\small marco.huber@ieee.org}%
}
}

\begin{document}
\begin{acronym}    
    \acro{FTS}{Force-Torque-Sensor}
    
    \acrodefplural{FTS}[FTSs]{Force-Torque-Sensors}
    \acro{LfD}{Learning-from-Demonstration}
    \acro{RL}{Reinforcement Learning}
    \acro{IL}{imitation learning}
    \acro{CNN}{convolutional neural network}
    \acro{MPC}{model predictive control}
    \acro{MSE}{mean squared error}
    \acro{ACT}{Action Chunking with Transformers}
    \acro{BC}{behavioral cloning}
    \acro{NN}{neural network}
    \acrodefplural{DSL}[DSLs]{Domain Specific Languages}
\end{acronym}

\maketitle
\thispagestyle{empty}
\pagestyle{empty}

\begin{abstract}

Automating the assembly of wire harnesses is challenging in automotive, electrical cabinet, and aircraft production, particularly due to deformable cables and a high variance in connector geometries. In addition, connectors must be inserted with limited force to avoid damage, while their poses can vary significantly. While humans can do this task intuitively by combining visual and haptic feedback, programming an industrial robot for such a task in an adaptable manner remains difficult. 
This work presents an empirical study investigating the suitability of behavioral cloning for learning an action prediction model for connector insertion that fuses force–torque sensing with a fixed position camera. We compare several network architectures and other design choices using a dataset of up to 300 successful human demonstrations collected via teleoperation of a UR5e robot with a SpaceMouse under varying connector poses.
The resulting system is then evaluated against five different connector geometries under varying connector poses, achieving an overall insertion success rate of over 90\,\%. 
 
\end{abstract}

\section{Introduction}
\label{sec:Introduction}
The assembly of wire harnesses is an important step in vehicle production, yet it remains
largely manual despite advances in industrial automation. Manual assembly is
labor-intensive and prone to errors, motivating automation~\cite{NavasReascos2022}.  
Automation is challenging because wire harnesses are deformable with a high variety of connector types and harness configurations. Moreover, connector installation involves contact-rich manipulation under tight tolerances~\cite{Nguyen2021}.
Connector installation consists of three steps: 
\begin{enumerate*}[label=(\roman*)]
\item inserting one connector into its counterpart, 
\item locking the connection, and \item verifying proper electrical contact. 
\end{enumerate*}
This work addresses connector insertion, i.e., the first step, where a movable harness-side connector (oftentimes called \emph{plug}), held by hand or a robotic gripper, is inserted into a fixed device-side connector (\emph{socket}), such as in a car door. 

Current robotic approaches for connector insertion often rely on rule-based search strategies, such as stride or spiral searches~\cite{ERF_2025, jiang_state_2022}. These methods depend on process parameters defined by human experts. Parameter tuning is typically iterative and time-consuming, which restricts the transferability of such approaches.  
In contrast, human operators carry out connector insertion reliably using visual and haptic coordination. This suggests an opportunity for data-driven methods that learn from demonstrations and imitate human strategies. Such methods could reduce the dependence on manual parameter tuning and improve adaptability to new connector types.

A common approach is to pretrain a deep learning model using \ac{IL} and then fine-tune it with
\ac{RL} to improve performance~\cite{Ankile2025, Rajeswaran2017, Wagenmaker2025}. In \ac{IL}, and in particular \ac{BC}, a policy is trained to mimic expert actions, providing a useful initialization for later \ac{RL} fine-tuning. However, successful pretraining requires that the model already achieves a reasonable baseline success rate. 
This leads to the research question of this work: \textit{What connector insertion success rates can deep learning models trained with \ac{BC} achieve?}
Although \ac{BC}, is a wide research field, there is a lack of investigations in the application of connector insertion w.r.t. suitable \ac{NN} architectures, number of sufficient demonstrations, and application capabilities of such models to different connector geometries and tolerances. 
\begin{figure}[t]
    \centering
    \vspace*{0.17cm}
    \includegraphics[width=0.48\textwidth]{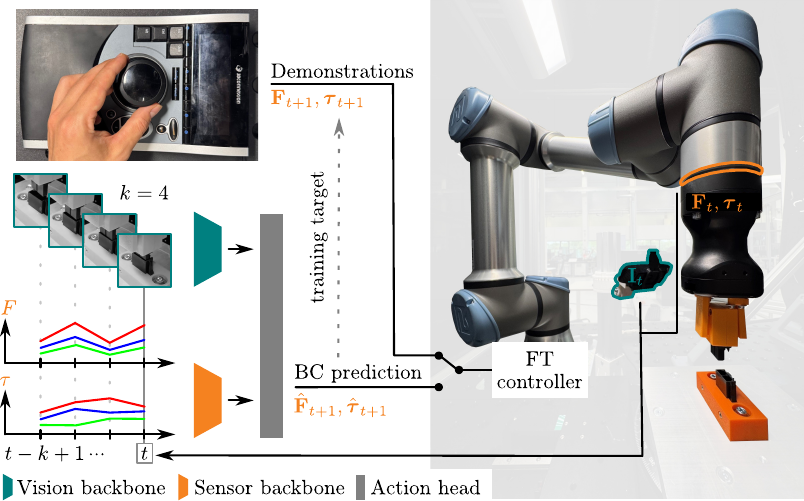}
    \caption{\ac{BC} predictor based on human demonstrations. Force–torque controller targets are generated using a space mouse and a \ac{NN} is trained to imitate insertion strategies from camera and force–torque sensor data.}
    \label{fig:introduction_system_concept}
    \vspace*{-0.7cm}
\end{figure}
This paper makes the following contributions:
\begin{itemize}
    \item An application of \ac{BC} for connector insertion trained from human demonstrations, as illustrated in Fig.~\ref{fig:introduction_system_concept}.
    \item An empirical analysis that shows that force–torque sensing combined with a grayscale monocular camera is sufficient for connector insertion with an average success rate of over \SI{90}{\percent}.
    \item An ablation study of several multi-modal NN architectures, composed of off-the-shelf network components like ResNets or Vision Transformers. 
    \item An experimental validation on a range of five automotive connectors with a pose tolerance of up to \SI{20}{\mm} and \SI{10}{\degree}.  
\end{itemize}

\section{Related work}
\label{sec:Related_work}
This section reviews tolerances in connector assembly, rule-based insertion
strategies, and imitation learning, with a focus on connector insertion.

\textbf{Rule-based Search Strategies}
\label{sub_sec:Rule-based_search_strategies}
are viable when connector poses are known with high precision and a
pose-controlled planner can compute an insertion path. In practice, however, exact pose estimation is limited by part tolerances, assembly tolerances, and sensor noise.
Reported tolerances in the literature vary depending on the connector type and task.  
For VGA connectors, pose estimation accuracies of \SI{0.6}{\mm} in translation and \SI{0.4}{\degree} in rotation have been achieved~\cite{Yu2019}. Tolerances of
\SI{\pm 1}{\mm} and \SI{2}{\degree} are assumed for peg-in-hole tasks~\cite{Gubbi2020}, while for automotive connectors \SI{\pm 2}{\mm} have been reported~\cite{kienle2025}. Other studies describe maximum tolerances of up to \SI{1.68}{\mm} for automotive connector installation~\cite{Yumbla2020} or as low as \SI{0.46}{\mm} and \SI{0.94}{\degree}, for automotive clips~\cite{Wnuk2023}.
To compensate for these uncertainties, rule-based search strategies are used. They define a sequence of parameterized movements together with termination conditions, enabling the robot to search for a feasible insertion path. Typical examples include spiral and spike search patterns.
In prior work, we evaluated a rule-based approach using a parameterized stride search with additional wiggling motions to reduce friction, achieving an overall success rate of \SI{76.7}{\percent}~\cite{ERF_2025}.
A central challenge of rule-based strategies is choosing suitable parameters for the search movements. In practice, engineers typically choose an initial parameter set from prior experience and refine it iteratively. Recent work seeks to automate this process by initializing with expert-chosen parameters and then optimizing using a data-driven approach~\cite{kienle2025}.

\label{sub_sec:related_work}
\textbf{\Acf{IL}} offers an alternative to rule-based approaches by enabling robots to learn search strategies directly from human demonstrations. The search is represented as a sequence
of steps, where at each step $t$ an action, such as a control command, is predicted
from an observation, e.g., an image or sensor reading. Such models can fuse multiple
modalities, for example RGB images with force--torque and pose data~\cite{liu2025}. Likewise, different action representations have been explored, including absolute end-effector poses~\cite{Ke2021}, incremental pose changes~\cite{Brohan2022}, or target wrenches for force control~\cite{liu2025}.  
A common approach to \ac{IL} is \ac{BC}, which treats \ac{IL} as a supervised learning problem based on demonstrations. 
The network architecture depends on the observation modality. For image inputs, pretrained convolutional networks, like ResNet or EfficientNet~\cite{Brohan2022}, are employed, while sensor data can be processed with feedforward networks~\cite{Ke2021}.
The training performance can be improved when the network receives a history of observations with $k$ previous steps, rather than single-step inputs~\cite{Brohan2022}. 
\ac{BC} has been applied to peg-in-hole tasks as an abstraction of connector insertion~\cite{Gubbi2020, Zang2023}.
In those approaches, the insertion targets were fixed in place, and pose data served as input to the \ac{BC} models. 
A common criticism is that using pose data may lead the models to simply memorize connector poses, thereby bypassing visual or haptic search strategies.
Even if \ac{BC} shows some interesting results, it suffers from compounding errors: Small deviations accumulate in the early steps, causing the system to leave the training distribution and enter unrecoverable states~\cite{Zhao2023}.
Several strategies have been proposed to mitigate this issue. One approach is to train models to output a sequence of $h$ future actions instead of a single step. At execution time, only the first action is applied, similar to \ac{MPC}. Other methods are reviewed in~\cite{Zhao2023}.
Another strategy is to combine \ac{BC} and \ac{RL}. A policy is first pretrained
with \ac{BC} on expert data and then fine-tuned with \ac{RL} to allow adaptation under distribution shifts. Such pretraining has been applied successfully to robotic manipulation tasks~\cite{Rajeswaran2017, Wagenmaker2025, Ankile2025}.
Recent advancements in large transformer architectures show promising results like ALOHA~\cite{Zhao2023}. It combines teleoperated demonstrations with a transformer-based model that predicts action chunks over a prediction horizon. This enables tasks such as battery insertion with \SIrange{80}{90}{\percent} success rate from only ten minutes of demonstrations.
RT-1 is another transformer-based \ac{IL} approach trained on 130,000 demonstrations covering over 700 tasks~\cite{Brohan2022}. It converts visual input using an EfficientNet backbone and natural language instructions into tokens and outputs discretized robot actions. In experiments, RT-1 achieved a \SI{97}{\percent} success rate on seen tasks and \SI{76}{\percent} on unseen ones. 
Vision-Language-Action models have also been applied to connector insertion, achieving \SI{25}{\percent} success on USB insertion and \SI{80}{\percent} success on electrical plug insertion~\cite{yu2025forcevla}.
For the scope of this work \ac{BC} pretraining and \ac{RL} fine-tuning seems like an interesting approach. We therefore focus first on the pretraining stage. Although \ac{BC} has been widely studied, its application to connector insertion holds room for additional research. Open questions include the suitability of different network architectures, the number of demonstrations required, and the ability to be applicable across heterogeneous connector geometries. In addition, most validation for connector installation is performed in stationary scenarios with a fixed connector pose, raising questions about generalization to assembly tolerances. Moreover, end-to-end approaches for force-controlled systems without relying on pose data have not yet been widely demonstrated for connector installation.  

\section{Problem formulation}
\label{sec:Problem_formulation}
We consider connector insertion as a control problem executed by a force--torque controller of the form  
\begin{equation}
    \mathbf{u}_t = \mathbf{C}\!\left(\hat{\mathbf{w}}_t - \mathbf{w}^{\mathrm{meas}}_{t}\right)~,
\end{equation}  
where $\mathbf{u}_t$ denotes the control output, $\mathbf{C}(\cdot)$ the controller, $\hat{\mathbf{w}}_t$ the target wrench, and $\mathbf{w}^{\mathrm{meas}}_{t}$ the measured wrench at time step $t$. 
The wrench is defined as
$\mathbf{w}_t = (\mathbf{F}_t, \boldsymbol{\tau}_t)$,
with forces $\mathbf{F}_t = (F_{t,x}, F_{t,y}, F_{t,z})$ and torques $\boldsymbol{\tau}_t = (\tau_{t,x}, \tau_{t,y}, \tau_{t,z})$.  
Accordingly, the target wrench is given by  
$\hat{\mathbf{w}}_t = (\hat{\mathbf{F}}_t, \hat{\boldsymbol{\tau}}_t)$
where $\hat{\mathbf{F}}_t$ and $\hat{\boldsymbol{\tau}}_t$ denote the target forces and torques, respectively. 
The connector insertion process can be interpreted as a sequence of target wrenches 
$\hat{\mathbf{w}}_1, \hat{\mathbf{w}}_2, \dots, \hat{\mathbf{w}}_{T},$  
for the controller. The goal is to learn a model $f_\theta$ that predicts the target wrench for the next time step $t+1$ based on the history of the previous $k$ steps, where $k>0$ is a hyperparameter. Setting $k=1$ corresponds to using only the current time step, while $k=10$ corresponds to using the current step together with the nine preceding steps, i.e., a history of ten steps in total. Formally,  
\begin{equation}
\hat{\mathbf{w}}_{t+1} = f_\theta(\mathbf{w}_{t-k+1:t}~, \mathbf{I}_{t-k+1:t})~,
\end{equation} 
where $\mathbf{w}_{t-k+1:t}$ denotes a history of measured wrenches and $\mathbf{I}_{t-k+1:t}$ a history of camera images from time steps $t-k+1$ to $t$. 
The predicted target wrench $\hat{\mathbf{w}}_{t+1}$ can then be applied by a force--torque controller $\mathbf{C}(\cdot)$.
For early time steps where $t < k$, the missing history entries in $\mathbf{w}_{t-k+1:t}$ and $\mathbf{I}_{t-k+1:t}$ are masked with zero values, i.e., $\mathbf{w}_{t<0} = \mathbf{0}$, $\mathbf{I}_{t<0} = \mathbf{0}$~.

For learning such a prediction model $f_\theta$, we assume a set of $M$ demonstrations as training data.  
A demonstration corresponds to manipulating the connector from an initial, non-inserted state to the final, inserted state. 
Each demonstration $m$ consists of $T_m$ time steps, where $T_m$ may vary between demonstrations.  
Thus, each demonstration provides a dataset of the form  
\begin{equation}
\mathcal{D}_{m} = \left\{ \mathbf{w}_t, \mathbf{I}_t\right\}_{t=1}^{T_m}~, \quad m = 1,\dots,M~,
\label{eq:dataset}
\end{equation}
where $\mathbf{I}_t$ denotes the image at time step $t$, and $\mathbf{w}_t$ denotes the measured wrench consisting of forces and torques in the task space of the end-effector.

We now frame the control problem into a \ac{BC} formulation.  
The observation at time step $t$ in demonstration $m$ is defined as  
$\mathbf{o}_{t,m} = \mathcal{D}^{t-k+1:t}_{m},$ 
i.e., the history of images and measured wrenches from time steps $t-k+1$ to $t$.  
The action is defined as the target wrench to be predicted, thus  
$\mathbf{a}_{t+1,m} = \hat{\mathbf{w}}_{t+1,m} =f_\theta(\mathbf{o}_{t,m}) \in \mathbb{R}^{d_{\text{action}}}$ with $d_{\text{action}}=6$.
The optimal parameters of the prediction model $f_\theta$ are obtained by solving the optimization problem  
\begin{equation}
\theta^\star  = \arg\min_\theta \sum_{m=1}^M \sum_{t=1}^{T_m-1} 
\big\| f_\theta(\mathbf{o}_{t,m}) - \mathbf{w}_{t+1,m} \big\|^2~.
\label{eq:theta_opt}
\end{equation} 
In other words, the model is trained by minimizing the \ac{MSE} between the predicted and the demonstrated target wrenches across all demonstrations.

While this single-step prediction suffices to imitate the demonstrations, prior work (see Sec.~\ref{sec:Related_work}) has shown that predicting multiple actions simultaneously can reduce compounding errors during model inference.  
Motivated by this, we extend the formulation such that $f_\theta$ outputs a sequence of predicted actions $\mathbf{a}_{t+1:t+h,m}$ over a horizon $h>0$:  
\begin{equation}
\theta^\star = \arg\min_\theta
\sum_{m=1}^M \sum_{t=1}^{T_m-h}
\sum_{i=1}^{h}
\big\|
f_\theta(\mathbf{o}_{t,m})[i]
- \hat{\mathbf{w}}_{t+i,m}
\big\|^2~,
\label{eq:theta_opt_h}
\end{equation}
where $[i]$ selects the $i$-th element of the predicted sequence.  
For $h=1$, this formulation reduces to the single-step case in~\eqref{eq:theta_opt}, while larger horizons ($h>1$) result in the prediction of longer action sequences.
At inference time, only the first predicted action is executed by the controller and then, the action sequence is predicted again, similar to \ac{MPC}. 

A connector insertion is considered successful, if $\| p_{\mathrm{goal}} - p_{\mathrm{meas}}\| < d$ holds, where $p_{\mathrm{goal}}$ is the goal pose of the end-effector, $p_{\mathrm{meas}}$ the actual pose of the end-effector, and $d$ the minimal distance where electrical contact is established.
Without loss of generality, the same criterion can be formulated with respect to other reference points on the robot (e.g., the connector), if there exists a fixed transform to $p_{\mathrm{meas}}$.

This work focuses on connector insertion across different geometries and does not explicitly consider electrical contact or securing of the locking latch, as including these steps would increase the experimental complexity.  
Therefore, in practice, we allow a slightly larger threshold $\tilde{d} > d$ for the evaluation of successful insertion, i.e., insertion is detected earlier than the establishment of electrical contact.  
Two assumptions motivate this focus: 
\begin{enumerate*}[label=(\roman*)]
\item the search strategy is the challenging part to automate with rule-based methods and 
\item the final press-fit for electrical contact can be accomplished with a force-controlled push, as shown in~\cite{ERF_2025}.
\end{enumerate*}

\section{Methodology}
\label{sec:Methodology}
The methodology describes the prediction model $f_\theta$, the hardware setup, and the data collection procedure.
\subsection{Imitation Learning}
\label{sub_sec:Imitation_Learning}
The \ac{IL} approach follows \ac{BC} and thus, is formulated as a supervised learning problem (see Sec.~\ref{sec:Problem_formulation}). The model $f_\theta$, represented by an \ac{NN} with parameters $\theta$, predicts the action corresponding to a target wrench. Training follows the optimization problem in \eqref{eq:theta_opt_h}. The hyperparameters, history length $k$ and prediction horizon $h$, are chosen empirically during evaluation. 
The network consists of two modality-specific backbones.
A vision backbone processes the visual input $\mathbf{I}_{t-k+1:t}$ and a sensor backbone processes time-series signals from the force--torque sensor $\mathbf{w}_{t-k+1:t}$. The features calculated by both backbones are concatenated into $\mathbf{x}\in\mathbb{R}^{d_\text{feature}}$, which is passed to an action head that outputs the predicted action sequence $\mathbf{a}_{t+1:t+h}$. A schematic of the network architecture is depicted in Fig.~\ref{fig:introduction_system_concept}. 
To select suitable backbones, we use classification networks under the hypothesis that their performance transfers to \ac{BC}.

\textbf{The Vision Backbone} consists of pretrained image or video classification networks that are employed from \ac{CNN} and Transformer families (see Tab.~\ref{tab:image_models}). PyTorch offers models and weights for image and video classification models pre-trained on ImageNet or Kinetics-400\footnote{Available at: \url{https://docs.pytorch.org/vision/main/models.html}}. Although these models expect RGB images of size $224\times224$, stacked grayscale images are used instead. Prior work showed that grayscale inputs can improve convergence in data-driven methods, at the cost of losing color information~\cite{ISR2023}, which we assume to be irrelevant here. The first convolutional layer is adapted to match the number of grayscale channels, requiring re-training of its weights. 

\textbf{For the Sensor Backbone}, we evaluate the following architectures, for which no pretrained weights are available: 

\begin{enumerate}[label=\arabic*)]
    \item A feedforward network with three hidden layers with 128/128/64 neurons.  
    \item An LSTM network with three layers, each with hidden size 128.  
    \item A 1D-CNN with three convolutional layers: 64 filters with kernel length $5$, 128 filters with length $5$, and 64 filters with length $3$, followed by adaptive average pooling.  
    \item A ROCKET model with 512 random kernels of length $\{1,3,5\}$ and dilations up to 2, producing 1024 features projected to 128 dimensions~\cite{Dempster2020}.  
\end{enumerate}
The sensor backbone architectures are parameterized to have a roughly comparable number of trainable parameters.  

\textbf{The Action Head} is evaluated with three design choices:  
1) a feedforward network with two hidden layers of 256/128 neurons,  
2) a transformer encoder, and  
3) a full transformer.  
The feedforward network maps the feature vector $\mathbf{x} \in \mathbb{R}^{d_\text{feature}}$ to $a_{t+1:t+h} \in \mathbb{R}^{h\times d_\text{action}}$.  
The transformer encoder first projects the input feature vector $\mathbf{x}\in\mathbb{R}^{d_\text{feature}}$ into the model width $d_1$ using a linear layer, resulting in $\mathbf{z}\in\mathbb{R}^{d_1}$ and replicates it across the prediction horizon with a learned temporal embedding $\mathbf{E}$ according to
$
\mathbf{T} =
\big[\;\mathbf{z} \; \cdots \; \mathbf{z}\;\big]^\top + \mathbf{E}
\in \mathbb{R}^{h\times d_1}.
$
The sequence $\mathbf{T}$ is processed by a transformer encoder with $L=2$ layers, $n_\text{head}=4$ attention heads, and model width $d_2=128$, producing an output sequence $\mathbf{Y}\in\mathbb{R}^{h\times d_2}$. Finally, a second linear layer projects each of the $h$ steps in $\mathbf{Y}$ to the predicted sequence $a_{t+1:t+h}\in\mathbb{R}^{h\times d_\text{action}}$. 
The full transformer uses an encoder–decoder formulation. The input feature vector $\mathbf{x}\in\mathbb{R}^{d_\text{feature}}$ is projected into the model width $d_3=64$ and treated as a source sequence $\mathbf{S} \in \mathbb{R}^{1\times d_3}$ of length $1$. The decoder input consists of $h$ learned query embeddings $\mathbf{Q} \in \mathbb{R}^{h\times d_3}$, where each query corresponds to one step of the prediction horizon. The transformer processes the pair $(\mathbf{S}, \mathbf{Q})$ with $l=2$ layers and $n_\text{head}=4$ attention heads, producing $\mathbf{Y} \in \mathbb{R}^{h\times d_3}$. Finally, a linear layer maps each of the $h$ decoder outputs to the action sequence $a_{t+1:t+h} \in\; \mathbb{R}^{h\times d_\text{action}}$.  
To enable a fair comparison, the action head networks are designed with similar parameter counts.  

\subsection{Hardware Setup}
\label{sub_sec:Hardware_setup}
\begin{figure}[t]
    \centering
    \vspace*{0.17cm}
    \subfloat[Car door connector setup. The socket is fixed and the robot holds the plug with start poses varied along the shown directions. The camera is visible in the background.]{     
        \centering
        \includegraphics[width=0.46\linewidth]{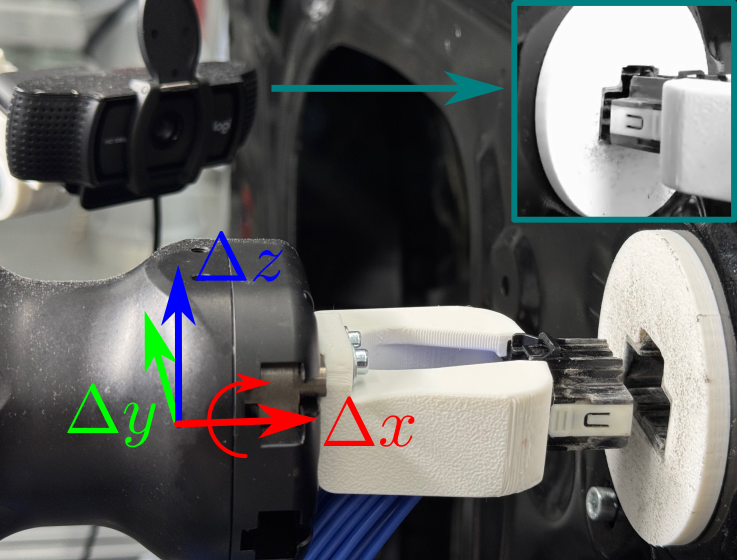}
        \label{fig:Hardware_setup:car_connector}
    }
    \hfill
    \vspace*{0.1cm}
    \subfloat[Setup for additional connector geometries. The socket is mounted on a three-axis positioning system with varied start poses, while the robot holds the plug at a fixed start pose.]{
            \centering
            \includegraphics[width=0.46\linewidth]{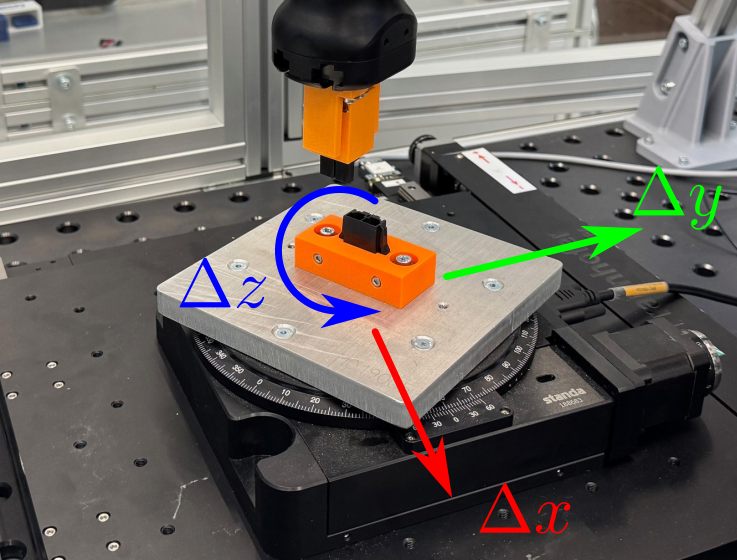}
            \label{fig:Hardware_setup:positioning_system}
    }
    \caption{Hardware setup used in the experiments.}
    \label{fig:Hardware_setup:hardware_setup}
    \vspace*{-0.7cm}
\end{figure}

The experiments are conducted with a Universal Robot UR5e equipped with a Robotiq Hand-E gripper.
The available sensors are the integrated force--torque sensor in the robot flange and a Logitech webcam with VGA resolution that is pointed at the connector, see Fig.~\ref{fig:Hardware_setup:car_connector}. The webcam is used because the vision backbones operate on $224\times224$ grayscale images, so higher resolution cameras would be downsampled while adding computational cost. The camera and lighting are assumed to be static, as these conditions can be controlled in industrial environments.
The robot control is executed as a ROS2 node using the open-source Cartesian force controller from FZI with their default gains for the UR5e~\cite{FDCC}. 
The robot is assumed to already hold the plug, as picking is not part of the examined problem. Successful wire harness picking has been demonstrated by Zürn et al.~\cite{Zuern2025}. 
The gripper fingers are 3D-printed to achieve a form-fit with the connector.
Directly gripping the cable is excluded, as some automotive OEM guidelines prohibit this practice due to wire damage risks.
Human demonstrations are recorded using a 3Dconnexion SpaceMouse, providing a six degree of freedom teleoperation and support by open-source drivers\footnote{Available at: \url{https://wiki.ros.org/spacenav_node}}.

The insertion of the lock connector in a car door is demonstrated as shown in Fig.~\ref{fig:Hardware_setup:car_connector}. The socket is mounted stationary on a flat surface of the door. A Molex Stac64 20-pin connector is used as a substitute for the original connector with similar shape and size. 
The plug can only be inserted to about \SI{80}{\percent} before the gripper fingers collide with the door, which requires a regrasp. A successful insertion is therefore defined as reaching the threshold $\tilde{d}$, corresponding to \SI{80}{\percent} of the full insertion depth $d$.

To simulate assembly tolerances, start poses for the end effector are sampled from a region around the socket with a range of \SI{20}{\mm} in translation and \SI{10}{\degree} about the $x$-axis (insertion axis), see Fig.~\ref{fig:Hardware_setup:car_connector}. Other rotations are not considered, as compensation along the insertion axis is the most intuitive for humans and sufficient as a proof of concept. The selected tolerances correspond to roughly \SI{50}{\percent} of the connector width and are about one order of magnitude larger than typically addressed by rule-based systems, which operate in the range of \SI{2}{\mm}. The proposed approach is designed for larger uncertainties where precise positioning or sensor systems are not available. 
To evaluate the approach on additional geometries, a second experiment is conducted with four connectors mounted on a three-axis positioning system. After each insertion trial, the connector is shifted to simulate tolerances, while the robot’s starting pose remains fixed. A tolerance region include translations in $x$ and $y$ by again \SI{50}{\percent} of the connector width and rotations of \SI{10}{\degree} around the $z$-axis, as shown in Fig.~\ref{fig:Hardware_setup:positioning_system}.

\subsection{Data Collection}
\label{sec:methodology_data}
To improve robustness against varying lighting conditions, image augmentation is applied during training. Brightness is adjusted by shifting the mean intensity of the grayscale image by up to \SI{20}{\percent}, and contrast is increased by scaling the standard deviation by up to \SI{20}{\percent}. The augmentations are applied randomly to the entire image history during training. 
All sensory inputs and actions are normalized. Observations are $z$-normalized using the empirical mean and standard deviation computed from the demonstrations. The action space is scaled to $[-1,1]$, where $1$ corresponds to a motion along or around a coordinate axis and $-1$ motion in the opposite direction.  
During teleoperation, the operator inserts the connector by performing translational and rotational motions. If the connector jams on an edge, the operator applies corrective motions. In addition, the output is overlayed with sinusoidal oscillations as in~\cite{ERF_2025} to reduce friction. The control frequency for both demonstration collection and model inference is fixed at \SI{20}{\hertz} due to software limitations. This value is a conservative estimate of the average software execution time, including Python overhead, data logging, disk writing and model inference. We hypothesize that increasing the control frequency would not substantially improve success rate, as human operators are able to perform robust insertion with this control frequency. Related work reports reduced execution time with higher frequencies, but not higher success rates~\cite{Zhao2023}.

We define three datasets:
The training set consisting of $M$ human demonstrations generated from an equidistant grid of insertion start poses. The translational range of \SI{50}{\percent} of the connector width in $x,y,z$ is discretized into $m_x, m_y, m_z$ points per axis, resulting in $M = m_x m_y m_z$ equidistant starting positions. The rotational range of \SI{10}{\degree} is discretized into $M$ equidistant rotations. Rotations and positions are sampled independently to form the start poses for human demonstrations. 
Initially, the grid is chosen to yield approximately 100 demonstrations, while in Sec.~\ref{sec:Evaluation} the effect of the number of demonstrations on the success rate is investigated.
The validation set also consists of human demonstrations. From the same tolerance region, $N$ start poses are sampled uniformly at random to measure prediction accuracy. Models are trained for ten epochs and evaluated after each epoch. The final checkpoint is selected by lowest validation \ac{MSE}. 
For testing, a separate set of randomized start poses is generated. The selected model is tested on the real robot using $J$ uniformly sampled, unseen start poses from the tolerance region.

\section{Evaluation}
\label{sec:Evaluation}
The first experiments aim to identify a suitable network architecture and to investigate the influence of the history length $k$, prediction horizon $h$, and the required amount of demonstration data. These experiments are conducted on the car door connector shown in Fig.~\ref{fig:Hardware_setup:car_connector}. After selecting an architecture and parameters, the system performance is further evaluated on different connector geometries using the three-axis positioning system, shown in Fig.~\ref{fig:Hardware_setup:positioning_system}.  
In all experiments, performance is measured by the task success rate, defined as the ratio of successful insertions to total trials. A trial is successful if the connector reaches the maximum insertion depth $\tilde{d}$ (cf. Sec.~\ref{sec:Problem_formulation}) within the time limit of 300 steps. This limit is based on human demonstrations, which require about 120 steps on average (\SI{\approx 6}{\second}), and up to 250 steps (\SI{\approx  12.5}{\second}) in difficult cases with connector jamming. 

\subsection{Car Door Connector}
\label{sub_sec:Car_door_connector}
For the car door connector a tolerance grid with $m_x=3$, $m_y=6$, and $m_z=6$ is generated, resulting in $M=108$ demonstrations as the training dataset. The discretization along $x$ (insertion axis) is coarser, as motion is mainly forward, while higher resolution is used in $y$ and $z$ for the search motion. Data collection takes \SI{22}{\minute}, including insertions and resetting to new start poses.
The validation set comprises $N=10$ demonstrations with randomized starting poses sampled from the tolerance region. For testing, the system is evaluated on $J=30$ trials with unseen start poses. 
All networks are trained with the same learning rate ($1 \mathrm{e}{-3}$), batch size (8), and number of epochs (10). The input history is fixed for the first experiments to $k=5$ and the prediction horizon to $h=1$. Each network is trained and evaluated three times using different random seeds. Different seeds are used to average over stochastic training and inference dynamics. 

\textbf{Vision Backbones} are evaluated with different pretrained networks, while the sensor backbone and action head are fixed as a feedforward network. The results are reported as mean $\mu_{\mathrm{SR}}$ and standard deviation $\sigma_{\mathrm{SR}}$ of the success rate over three random seeds resulting in 90 test trials for each network architecture (see Tab.~\ref{tab:image_models}). This experiment uses the largest number of trials, based on the assumption that the vision backbone has the strongest impact on success rate due to its higher parameter count.

\begin{table}[t]
\vspace*{0.17cm}
\caption{Comparison of image backbones by success rate ($\mu_{\mathrm{SR}}$, $\sigma_{\mathrm{SR}}$), \ac{MSE}, total number of model parameters $n_\theta$ (in millions) and worst-case inference time $t_{\text{Inf},\max}$ (in milliseconds).}
\centering
\begin{tabular}{l c c c c c}
\hline
\textbf{Model} & $\bm{\mu}_{\mathrm{SR}}$ & $\bm{\sigma}_{\mathrm{SR}}$ & \textbf{MSE} & $\bm{n}_{\theta}$ & $\bm{t}_{\text{Inf, max}}$ \\
\hline
RegNetx3\_2GF & \textbf{93.3} & 5.80 & 0.0376 & 14.6 & 3.14 \\
DenseNet201 & 91.1 & 15.4 & 0.0373 & 18.7 & 6.04 \\
DenseNet121 & 90.0 & 12.0 & 0.0370 & 7.3  & 4.20 \\
EfficientNetV2S & 84.4 & 24.1 & 0.0372 & 20.6 & 4.45 \\
RegNety8GF & 81.1 & 27.1 & 0.0374 & 38.0 & 2.74 \\
V-ResNetR3D & 80.0 & 31.8 & 0.0384 & 33.4 & 5.99 \\
DenseNet169 & 80.0 & 11.5 & 0.0371 & 13.0 & 5.08 \\
EfficientNetV2M & 78.9 & 25.2 & 0.0373 & 53.3 & 6.05 \\
V-ResNetMC3 & 76.7 & 37.6 & 0.0390 & 11.7 & 5.73 \\
RegNetx1\_6GF & 76.7 & 40.4 & 0.0370 & 8.6  & 2.31 \\
ResNet50 & 76.7 & 14.5 & \textbf{0.0369} & 24.1 & 1.73 \\
V-ResNetR2P1D & 74.4 & 31.0 & 0.0376 & 31.5 & 10.3 \\
RegNetx8GF & 73.3 & 28.5 & 0.0376 & 38.2 & 2.50 \\
RegNetx16GF & 65.6 & 33.7 & 0.0386 & 52.8 & 3.25 \\
RegNety3\_2GF & 62.2 & 34.0 & 0.0376 & 18.4 & 3.26 \\
ResNet34 & 60.0 & 36.1 & 0.0373 & 21.5 & 1.40 \\
ResNet18 & 60.0 & 38.4 & 0.0375 & 11.4 & 0.84 \\
RegNety1\_6GF & 51.1 & 30.1 & 0.0382 & 10.6 & 4.42\\
RegNety32GF & 48.9 & 36.0 & 0.0383 & 142 & 6.78 \\
RegNetx32GF & 26.7 & 3.30 & 0.0379 & 106 & 6.25 \\
RegNety16GF & 25.6 & 3.80 & 0.0386 & 81.4 & 4.05 \\
ViT\_B\_32 & 13.3 & 5.80 & 0.0437 & 89.3 & 1.99 \\
SwinT\_T & 1.10 & 1.90 & 0.0487 & 27.8 & 3.54 \\
ConvNeXtTiny & 1.10 & 1.90 & 0.0484 & 28.1 & 1.73 \\
Swin3D\_t & 0.00 & 0.00 & 0.0485 & 28.1 & 6.45 \\
\hline
\end{tabular}
\label{tab:image_models}
\vspace*{-0.65cm}
\end{table}
Tab.~\ref{tab:image_models} reports success rate and the lowest \ac{MSE} achieved for each architecture across seeds, along with the total number of model parameters. The parameter count is included to examine potential correlations with success rate. In addition, worst-case inference times are reported to verify compliance with the \SI{20}{\hertz} control requirement.
Among the evaluated models, \texttt{RegNetx3\_2GF} achieves the highest mean success rate of \SI{93.3}{\percent}. However, success rates vary widely across RegNet variants, and there is no theoretical explanation for why this particular variant performs best. 
In contrast, the three transformer architectures are not suitable for this task, all reaching success rates of maximum \SI{13.3}{\percent}. A possible reason is that the dataset size is not large enough for transformers. 
The best-performing video classification model, \texttt{Video-ResNetR3D}, achieves a \SI{3.3}{\percent} higher success rate than its best-performing image classification \texttt{ResNet50} counterpart. However, five other image classification models still outperform the video model. The \texttt{Swin3D\_t} video model, in contrast, consistently yields a success rate of \SI{0}{\percent}.
Several models show large variance across seeds, with standard deviations up to \SI{40}{\percent}, indicating strong sensitivity to stochastic training and inference effects. \texttt{RegNetx3\_2GF} is more stable, with a standard deviation of \SI{5.8}{\percent}.
One noteworthy observation is that a low MSE does not necessarily correspond to a high success rate. Therefore, MSE during training cannot be used as a reliable indicator of success rate and experiments on the real system remain necessary. The top three networks have less than 20 million parameters, yet others of similar size perform worse, indicating no consistent relation to model size. Regarding inference time, all networks satisfy the \SI{20}{\hertz} time requirement. 
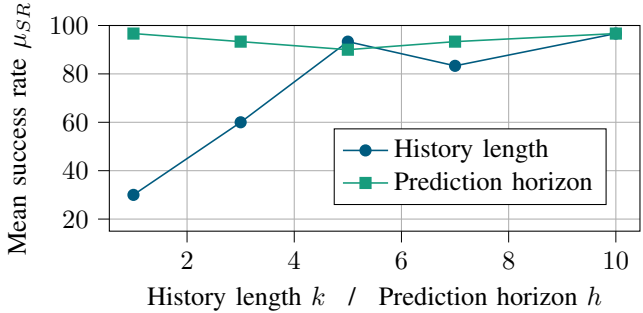
\begin{figure}[t]
    \centering
    \vspace*{0.17cm}




\begin{tikzpicture}

\definecolor{darkgray176}{RGB}{176,176,176}
\definecolor{steelblue31119180}{RGB}{0,91,127}
\definecolor{fraunhofergreen}{RGB}{23,156,125}

\begin{axis}[
    width=\linewidth,
    height=0.5\linewidth,
    tick align=outside,
    tick pos=left,
    x grid style={darkgray176},
    grid=both,
    xlabel={History length $k$ \, / \, Prediction horizon $h$},
    xmin=0.55, xmax=10.45,
    xtick style={color=black},
    y grid style={darkgray176},
    ylabel={Mean success rate $\mu_{SR}$},
    ymin=15, ymax=100,
    ytick style={color=black},
    legend style={at={(0.68,0.5)}, anchor=north, legend columns=1},
    legend cell align={left}
]

\addplot [semithick, steelblue31119180, mark=*, mark size=2]
table {%
1 30.000000000000
3 60.000000000000
5 93.3333333333335
7 83.33333333333337
10 96.6666666666667
};
\addlegendentry{History length}

\addplot [semithick, fraunhofergreen, mark=square*, mark size=2]
table {%
1 96.6666666666667
3 93.3333333333337
5 90.0000000000000
7 93.3333333333337
10 96.6666666666667
};
\addlegendentry{Prediction horizon}

\end{axis}
\end{tikzpicture}
    \caption{Mean success rate over different history lengths $k$ and prediction horizons $h$ on a shared $x$-axis. In blue different $k$ given $h=1$. In green different $h$ given $k=10$. }
    \label{fig:multistep_prediction}
    \vspace*{-0.7cm}
\end{figure}

Next, we study the effect of history length $k$ and prediction horizon $h$. The best-performing model, \texttt{RegNetx3\_2GF}, is used as the vision backbone. First, a grid search is performed over different history lengths $k \in [1,10]$ with a fixed prediction horizon $h=1$. For the value of $k$ with the highest mean success rate $\mu_{\mathrm{SR}}$, a second grid search is then carried out over prediction horizons $h \in [1,10]$. Every model is evaluated with three seeds on $J=30$ unseen start poses.
The blue curve in Fig.~\ref{fig:multistep_prediction} shows the relation between
$k$ and $\mu_{\mathrm{SR}}$. A positive correlation can be observed: Longer histories lead to higher success rates. For $k=1$, the robot often makes initial contact but then remains stationary until timeout, resulting in a low success rate. A possible reason is that the network learns the mean of a search path, which without temporal context is zero. For the same reason, wiggling motions are not reproduced. Wiggling corresponds to a temporal signal $\sin(t)$, but without history, $t$ cannot be inferred and the motion cannot be executed. As $k$ increases, the success rate improves and also the wiggling motion is reproduced. Based on these results, $k=10$ is chosen for subsequent experiments.

The green curve shows the success rate over the prediction horizon $h$. Across all horizons, the success rate lies within a range of \SIrange{90.0}{96.7}{\percent}. The dip at $h=5$ is not systematic, as neighboring values recover, and is therefore attributed to noise. Although the literature recommends using larger horizons, the experiments show that similar results can be achieved with $h=1$ or $h=10$. We adopt $h=10$ for subsequent experiments in line with prior work.

\textbf{Four Sensor Backbones} for time series data are evaluated: A feedforward network, an LSTM, a 1D-CNN, and a ROCKET model (cf. Sec.~\ref{sub_sec:Imitation_Learning}). For a fair comparison, the parameter counts $n_{\theta}$ of all models are chosen in a comparable range. All models are trained with $k=10$ and $h=10$ using the \texttt{RegNetx3\_2GF} vision backbone and every model is evaluated with three seeds and $J=30$ test trials.
\begin{table}[t!]
  \vspace*{0.17cm}
  \centering
  \caption{Comparison of different sensor backbones.}
  \label{tab:time_backbone_analysis}
  \begin{tabular}{l c c c c}
    \hline
    \textbf{Model} & $\bm{\mu}_{\mathrm{SR}}$ & $\bm{\sigma}_{\mathrm{SR}}$ &\textbf{MSE} & $\bm{n}_{\theta}$ \\
    \hline
    Feedforward & \textbf{96.7} & 2.86 & 0.0342 & 14,637,468 \\
    LSTM        & 80.0 & 5.43 & \textbf{0.0333} & 14,955,100 \\
    ROCKET      & 80.0 & 16.3 & 0.0356 & 14,752,476 \\
    1DCNN       & 66.7 & 11.0 & 0.0334 & 14,672,604 \\
    \hline
  \end{tabular}

\end{table}
Tab.~\ref{tab:time_backbone_analysis} summarizes the results. The feedforward network achieves the highest success rate, while the LSTM attains the lowest \ac{MSE}. This again shows that a low \ac{MSE} does not imply a high success rate. The 1D-CNN yields a similar \ac{MSE} to the LSTM but a lower success rate. The ROCKET model matches the LSTM in success rate.

\textbf{For the Action Head} three variants are compared: a feedforward network, a transformer encoder, and a full transformer architecture. All models are trained using the \texttt{RegNetx3\_2GF} vision backbone and the feedforward sensor backbone and each is evaluated on $J=30$ start poses.  
\begin{table}[t]
  \centering
  \vspace*{0.17cm} 
  \caption{Comparison of different action heads.}
  \label{tab:action_head_analysis}
  \begin{tabular}{l c c c c}
    \hline
    \textbf{Model} & $\bm{\mu}_{\mathrm{SR}}$ & $\bm{\sigma}_{\mathrm{SR}}$ & \textbf{MSE} & $\bm{n}_{\theta}$ \\
    \hline
    Feedforward         & \textbf{96.7} & 2.86 & \textbf{0.0342} & 14,637,468 \\
    $\text{Transformer}_{\text{Full}}$     &   86.7 & 13.4 & 0.0388 & 14,748,966 \\
    $\text{Transformer}_{\text{Encoder}}$  & 80.0 & 6.65 & 0.0372 & 14,830,950 \\
    \hline
  \end{tabular}
  \vspace*{-0.7cm}
\end{table}
Tab.~\ref{tab:action_head_analysis} summarizes the results. The feedforward action head achieves the highest success rate and both  transformer models perform worse by at least \SI{10.0}{\percent}. 
Based on the conducted experiments the network architecture with the highest success rate consists of \texttt{RegNetx3\_2GF} as the vision backbone and feedforward networks as the sensor backbone and action head.

Next, we study how success rate depends on the number of demonstrations. A tolerance grid with $m_x=3$, $m_y=10$, and $m_z=10$ yields $M=300$ demonstrations, recorded in \SI{68}{\minute}. From this set, subsets of $m \in \{10, 25, 50, 75, 100, 200, 300\}$ demonstrations are sampled to train
models. Validation uses $N=30$ start poses, and testing uses $J=30$ unseen start poses.
\begin{figure}[t]
    \vspace*{0.17cm}
    \centering
\begin{tikzpicture}

\definecolor{darkgray176}{RGB}{176,176,176}
\definecolor{lightgray204}{RGB}{204,204,204}
\definecolor{steelblue31119180}{RGB}{0,91,127}
\definecolor{fraunhofergreen}{RGB}{23,156,125}

\begin{axis}[
height=0.25\textwidth,
width=0.48\textwidth,
legend cell align={left},
legend style={
  fill opacity=0.8,
  draw opacity=1,
  text opacity=1,
  at={(0.97,0.03)},
  anchor=south east,
  draw=lightgray204
},
tick align=outside,
tick pos=left,
x grid style={darkgray176},
xlabel={Number of demonstrations $m$},
xmajorgrids,
xmin=-4.5, xmax=314.5,
xtick style={color=black},
y grid style={darkgray176},
ylabel={Mean success rate $\mu_{\mathrm{SR}}$},
ymajorgrids,
ymin=0, ymax=105,
ytick style={color=black}
]
\path [fill=steelblue31119180, fill opacity=0.12, line width=0pt]
(axis cs:10,30)
--(axis cs:10,0)
--(axis cs:30,20)
--(axis cs:50,50)
--(axis cs:70,50)
--(axis cs:100,70)
--(axis cs:200,83.3333333333333)
--(axis cs:300,100)
--(axis cs:300,100)
--(axis cs:300,100)
--(axis cs:200,100)
--(axis cs:100,100)
--(axis cs:70,93.3333333333333)
--(axis cs:50,83.3333333333333)
--(axis cs:30,66.6666666666667)
--(axis cs:10,30)
--cycle;
\addlegendimage{area legend, fill=steelblue31119180, fill opacity=0.12, line width=0pt}
\addlegendentry{Min--Max range}

\addplot [semithick, steelblue31119180, mark=*, mark size=3, mark options={solid}]
table {%
10 10
30 33.3333333333333
50 70
70 76.6666666666667
100 93.3333333333333
200 93.3333333333333
300 100
};
\addlegendentry{Median}

\addplot [fraunhofergreen, dashed, thick] 
coordinates {(-4.5,76.7) (314.5,76.7)};
\addlegendentry{Stride search~\cite{ERF_2025}}

\end{axis}

\end{tikzpicture}
    \caption{Mean success rate given a number of demonstrations $m$ with a comparison of a rule based stride search strategy from~\cite{ERF_2025}.}
    \label{fig:Car_door_connector:success_rate_vs_num_demonstrations}
    \vspace*{-0.6cm}
\end{figure}
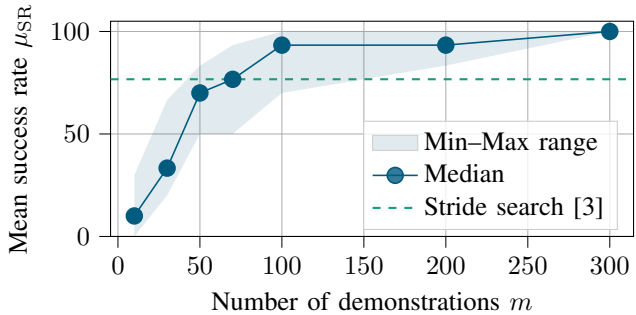
Fig.~\ref{fig:Car_door_connector:success_rate_vs_num_demonstrations} shows the success rate over the number of demonstrations. For clarity, the minimum, median, and maximum over three seeds are reported. A clear positive correlation is observed between success rate and the number of demonstrations, confirming that \ac{BC} benefits from more demonstrations.
A plateau effect is observed after 100 demonstrations, where models already achieve a success rate of \SI{100}{\percent} in some trials. 
However, the variance is larger compared to prior experiments with 108 demonstrations. With the full set of 300 demonstrations, the model converges to a success rate of \SI{100}{\percent} over all 90 evaluation trials. Thus, for subsequent experiments with different connector geometries, 300 demonstrations are collected. 
It should be noted that the observed \SI{100}{\percent} success rate refers to the tested 90 evaluation trials and deviations could have occurred if more trials were made.
In addition, we compare a conventional rule-based stride search from our prior work ~\cite{ERF_2025} on the same connector type with tolerances up to \SI{15}{\milli\meter}  (cf. the green dashed line in Fig.~\ref{fig:Car_door_connector:success_rate_vs_num_demonstrations}). After 200 demonstrations, the \ac{BC} approach surpasses the rule-based search regarding the success rate across all test cases.

\subsection{Other Connector Geometries}
\label{sub_sec:Other_connector_geometries}
We evaluate the proposed \ac{BC} method on additional connector geometries using the three-axis positioning system (Fig.~\ref{fig:Hardware_setup:positioning_system}). The socket is fixed in a 3D-printed holder. A tolerance grid with $m_x=3$, $m_y=10$ and $m_z=10$ is defined, resulting in $M=300$ demonstrations with an absolute tolerance of \SI{50}{\percent} of each connector width. The network configuration is a \texttt{RegNetx3\_2GF} vision backbone, a feedforward sensor backbone, and a feedforward action head. Training is performed on three seeds and each trained model is evaluated on $N=30$ demonstrations and tested on $J=30$  trials, resulting in 90 trials per connector. Performance is reported as mean $\mu_{\mathrm{SR}}$ and standard deviation $\sigma_{\mathrm{SR}}$ of the success rate.  
In addition, the execution speed of the \ac{BC} on the robot is compared to human performance on successful trials.  
Tab.~\ref{tab:connector_geometries} summarizes the results of these experiments.  

\begin{table}[t]
\centering
\vspace*{0.17cm} 
\caption{Evaluation on different connector geometries with success rate ($\mu_{\mathrm{SR}}, \sigma_{\mathrm{SR}}$), human/\ac{BC} insertion time in seconds (and steps), and translational tolerance in millimeter.}
\begin{tabular}{l c c c c }
\toprule
& \includegraphics[width=0.06\textwidth]{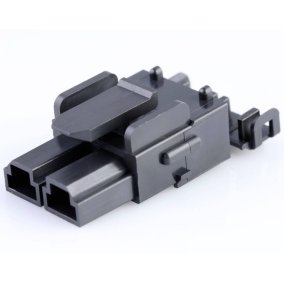} 
& \includegraphics[width=0.06\textwidth]{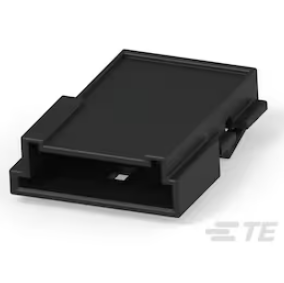} 
& \includegraphics[width=0.06\textwidth]{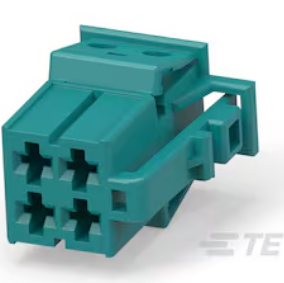} 
& \includegraphics[width=0.06\textwidth]{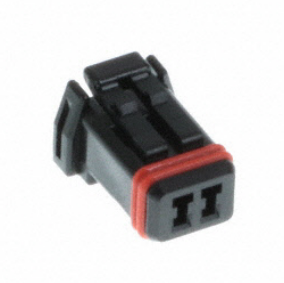} \\
& \includegraphics[width=0.06\textwidth]{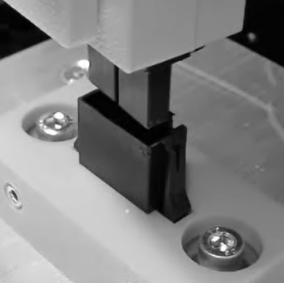} 
& \includegraphics[width=0.06\textwidth]{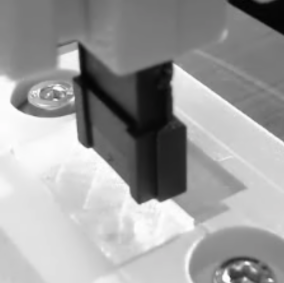} 
& \includegraphics[width=0.06\textwidth]{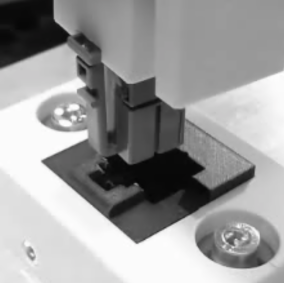} 
& \includegraphics[width=0.06\textwidth]{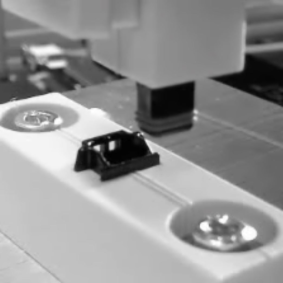} \\
\midrule
Manufacturer  & Molex & TE & TE & JAE \\
Series        & Mini-Fit & MQS & AMP MCP & MX19 \\
ID            & A & B & C & D \\
\midrule
$\mu_{\mathrm{SR}}$     & 93.3 & 94.4 & 90.0 & 93.3 \\
$\sigma_{\mathrm{SR}}$  & 5.47 & 2.01 & 8.10 & 5.47 \\
$t_{\text{human}}$     & 6.40 (128) & 9.45 (189) & 7.30 (146) & 7.20 (144) \\
$t_{\text{bc}}$        & 7.45 (149) & 9.55 (191) & 8.10 (162) & 7.90 (158) \\
Tolerance              & 10 & 10 & 8 & 4 \\
\bottomrule
\label{tab:connector_geometries}
\end{tabular}
\vspace*{-0.7cm}
\end{table}
Across all connectors, the method achieves a mean success rate of
\SI{92.8}{\percent}, showing applicability to different geometries. However, robustness varies across connectors: the standard deviation reaches \SI{8.10}{\percent} for Connector C but is only \SI{2.02}{\percent} for Connector B, suggesting that some geometries are more suitable for the proposed approach than others. 
Interestingly, Connector D achieves a success rate comparable to the others, despite having the smallest positional tolerance of \SI{4}{\milli\meter}. This corresponds to a $6.25\times$ higher demonstration density per square millimeter compared to the \SI{10}{\milli\meter} tolerance of Connectors A and B.
In general, data-driven approaches like \ac{BC} benefit from higher spatial demonstration density, as this broadens observation space coverage and reduces extrapolation during inference, improving robustness. However, here higher density does not yield improvements, suggesting inherent method limitations, such as compounding errors.

The insertion time of the \ac{BC} approach is on average \SI{9.55}{\percent} (or \SI{0.66}{\second}) slower than human demonstrations. 
Thanks to the force controller, no connectors were damaged in any trials. Typical failure modes include the plug getting stuck at an edge of the socket and the robot remaining in place, or, more rarely, the robot missing the connector entirely and moving past the socket without insertion.

\section{Discussion}
\label{sec:Discussion}
During the experiments we observed that a low evaluation MSE does not necessarily correlate with a high task success rate. One hypothesis for explaining this observation is that MSE averages errors over all states in the evaluation dataset, while failures are often caused by rare but critical states (e.g., jamming). These states have little influence on evaluation MSE but dominate the success metric. Therefore, direct evaluation via success rate remains necessary.

Moreover, the experiments showed that CNN-based vision backbones achieved success rates up to seven to eight times higher than transformer-based backbones. This indicates that, at least for the analyzed connector insertion task using \ac{BC}, CNNs perform considerably better than recent transformer architectures. A possible reason is the inductive bias of the CNNs combined with the limited dataset size. 

Compared to large-scale approaches such as Vision-Language-Action (VLA) models~\cite{yu2025forcevla}, RT-1~\cite{Brohan2022} or ALOHA~\cite{Zhao2023}, the proposed \ac{BC} approach achieves success rates in a similar range. These methods also report success rates below the reliability levels required for industrial deployment (e.g., ours \SI{92.8}{\percent}, USB connector insertion using VLA \SI{25}{\percent}~\cite{yu2025forcevla}, AA-battery insertion in ALOHA \SI{96}{\percent}~\cite{Zhao2023}, seen tasks in RT-1 \SI{97}{\percent}~\cite{Brohan2022}). Moreover, the \ac{BC} approach requires significantly fewer demonstrations (ours 300 vs. \mbox{RT-1} 130{,}000) to achieve success rates of over \SI{90}{\percent} in connector insertion. In addition, the smaller model size results in lower compute requirements, making the approach more practical for edge deployment, especially for real-time critical systems, which are required for force-control.
However, in terms of zero-shot or few-shot generalization, the proposed method cannot compete with large-scale models. ALOHA reports \SIrange{80}{90}{\percent} success with as few as 50 demonstrations, and models such as RT-1 achieve zero-shot generalization with a success rate of \SI{76}{\percent} across unseen tasks using natural language prompts. This highlights the trade-off between specialized approaches like ours, which rely on smaller models and limited data but address a specific task, and more general approaches that offer broader applicability at the cost of requiring large models and extensive datasets. 

\section{Conclusion}
\label{sec:Conclusion}
This work has demonstrated \ac{BC} for connector insertion with five different connector geometries, achieving an average success rate of \SI{92.8}{\percent}. Using only a monochrome grayscale camera and a force--torque sensor, the system is able to learn insertion strategies that combine visual and haptic data without requiring pose information. 
The collection of such demonstrations requires about one hour in the given setting, which is still acceptable when compared to our prior work with a manual rule-based search strategy setup and parameter tuning.
The evaluated tolerance ranges of the approach are up to $10$ times larger than those reported in related work, highlighting the potential to avoid costly high-precision pose estimation and instead rely on robust insertion strategies as presented here. 

Nevertheless, the proposed method does not consistently approach success rates of \SI{100}{\percent}, which is required for industrial automation. Therefore, it should be viewed as a viable pretraining step, as motivated in
Sec.~\ref{sec:Introduction}. In addition, per-connector demonstrations are required, so the method is not zero-shot applicable to unseen connectors.

Future work can investigate fine-tuning of the \ac{BC} models (e.g., by ~\ac{RL}) to further improve task success rate. In addition, the approach could be extended beyond individual connector geometries by investigating multi-connector or cross-geometry training setups.

\addtolength{\textheight}{-12cm}   





\section*{ACKNOWLEDGMENT}
This work has received funding from the Ministry of Science, Research and Arts of the Federal State of Baden-Wuerttemberg within the Innovations Campus Future Mobility (ICM) and the European Union's Horizon Europe Framework Program under \emph{euROBIN} - Grant No. 101070596. ChatGPT 5 was used to assist with language editing and grammar enhancement.


\bibliographystyle{IEEEtran}  
\balance
\bibliography{references}  

\end{document}